\documentclass[letterpaper, 10 pt, conference]{ieeeconf}

\IEEEoverridecommandlockouts

\usepackage{times}
\usepackage{microtype}
\usepackage{graphicx}
\usepackage{booktabs}
\usepackage{multirow}
\usepackage{amsmath,amssymb,amsfonts}
\usepackage{mathtools}
\usepackage{xspace}
\let\labelindent\undefined
\usepackage{enumitem}
\usepackage{url}
\usepackage[hidelinks]{hyperref}
\usepackage[capitalize,nameinlink]{cleveref}
\usepackage{algorithm}
\usepackage{algpseudocode}
\usepackage[table]{xcolor}
\usepackage{pifont}
\usepackage{caption}
\usepackage{adjustbox}
\newcommand{\method}{\textit{\textbf{\textsc{UniLoop}}}\xspace}
\newcommand{\tb}{\textbf}
\newcommand{\ul}{\underline}
\newcommand{\encoder}{\mathcal{E}}
\newcommand{\retriever}{\mathcal{A}}
\newcommand{\decoder}{\mathcal{D}}
\newcommand{\verifier}{\mathcal{V}}
\newcommand{\posehead}{\mathcal{P}}
\newcommand{\R}{\mathbb{R}}
\newcommand{\SE}{\mathrm{SE}}

\newcommand{\cmark}{\ding{51}}
\newcommand{\xmark}{\ding{55}}

\definecolor{bestresultcolor}{HTML}{A9D18E}
\definecolor{secondresultcolor}{HTML}{C6E0B4}
\definecolor{thirdresultcolor}{HTML}{E2F0D9}
\newcommand{\bestresult}[1]{\cellcolor{bestresultcolor}\textbf{#1}}
\newcommand{\secondresult}[1]{\cellcolor{secondresultcolor}#1}
\newcommand{\thirdresult}[1]{\cellcolor{thirdresultcolor}#1}
\DeclareFontFamily{U}{mathb}{}
\DeclareFontShape{U}{mathb}{m}{n}{
  <-5.5> mathb5
  <5.5-6.5> mathb6
  <6.5-7.5> mathb7
  <7.5-8.5> mathb8
  <8.5-9.5> mathb9
  <9.5-11.5> mathb10
  <11.5-> mathb12
}{}
\DeclareSymbolFont{mathb}{U}{mathb}{m}{n}
\DeclareMathSymbol{\drsh}{3}{mathb}{"EB}
\newcommand{\cornerarrow}{\raisebox{1pt}{$\drsh\ $}}

\begin{document}
\title{\LARGE \bf
\method: Unified Loop Closure Detection from Retrieval to
Geometric Verification and Relative Pose Estimation}

\author{Jingwen Yu$^{1,2}$, Pengyu Wang$^{1}$, Jin Wu$^3$, Weinan Chen$^4$, Ping Tan$^1$, and Hong Zhang$^2$
\thanks{$^1$CKS-Robotics Institute, Hong Kong University of Science and Technology, Hong Kong, China.}
\thanks{$^2$Department of Electronic and Electrical Engineering, Southern University of Science and Technology, Shenzhen, China.}
\thanks{$^3$School of Intelligent Science and Technology, University of Science and Technology Beijing, Beijing, China.}
\thanks{$^4$Biomimetics and
Intelligent Robotics Lab (BIRL), School of Electromechanical Engineering, Guangdong University of Technology, Guangzhou, China.}
\thanks{$^*$Jingwen Yu and Pengyu Wang contribute equally to this work. Corresponding author: Hong Zhang (hzhang@sustech.edu.cn)}
}

\makeatletter
\let\@oldmaketitle\@maketitle
\renewcommand{\@maketitle}{\@oldmaketitle
\centering
\vspace{0.3cm}
\captionsetup{type=figure, singlelinecheck=false,font=small}
\begin{tabular}{cccc}
\includegraphics[width=0.9\textwidth,trim=5mm 0mm 5mm 0mm]{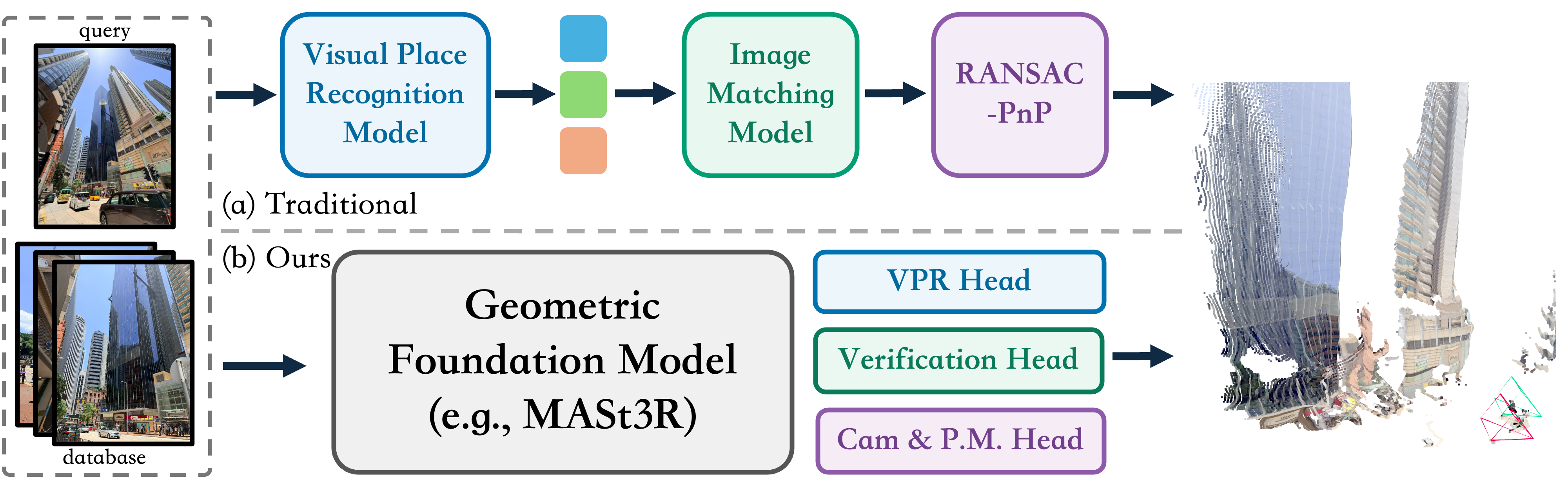}
\end{tabular}
\caption{\method unifies retrieval, geometric verification, and relative
pose estimation within a single MASt3R backbone. Unlike cascades of
separately trained visual models, \method reuses the image encoder for
global retrieval, the cross-view decoder for pairwise verification, and
the geometric heads for relative pose constraints, sharing an image
encoding across all stages. This framework proposes a new paradigm for loop closure detection, improving robustness and accuracy while reducing redundant computation.}
\label{fig:teaser}
\setcounter{figure}{1}
}
\makeatother
\maketitle

\begin{abstract}
Loop closure detection conventionally cascades separate models for global
retrieval, geometric verification, and relative pose estimation, leading to
redundant computation and disconnected visual representations. We propose
\method, a unified framework that reuses the hierarchical representations of
the pretrained geometric foundation model throughout this pipeline.
A shared image encoder produces features for compact global descriptors; the
same features are passed to the cross-view decoder for pairwise verification;
and the original geometric predictions provide relative camera constraints for
pose-graph optimization. Experiments on place recognition, loop verification,
and relative pose estimation show that \method improves loop-closure
robustness and accuracy while reducing redundant backbone computation relative
to modular pipelines. \method achieves the best Recall@10 on MSLS (96.6) and
Pitts30k (99.3) among the compared methods, while its verification branch
reaches 95.63 AP and 96.29 ROC-AUC, surpassing the existing geometric verification baselines.
\end{abstract}

\section{Introduction}
\label{sec:intro}

Loop closure detection lets a robot recognize previously visited
places, correct accumulated odometry drift, and build a globally
consistent map over long-term operation.
Despite its central role in Simultaneous Localization And Mapping
(SLAM), it is still commonly implemented as a multi-stage cascade of
separate visual models, as shown in Fig.~\ref{fig:teaser}.
A visual place recognition model retrieves candidate matches, a
geometric verification (e.g., local feature image matching model) module checks their consistency, and a relative pose module recovers
the camera pose between the query and its top candidate, which is
finally fed as a constraint into pose graph optimization.
Although effective, this design duplicates image encoding, prevents intermediate representation sharing, and yields independently optimized modules whose objectives are only loosely aligned with the final
loop closure decision.

Recent geometric foundation models (GFMs) offer a different opportunity.
MASt3R~\cite{leroy2024mast3r} contains a natural hierarchy of 3D-aware
visual representations, where its image encoder produces single-view features, its cross-view decoder constructs pair-conditioned representations, and its geometric heads predict dense 3D structure and matching cues.
This hierarchy aligns with the stages of loop closure detection: global
representations serve retrieval, while patch-level representations support
verification and relative pose estimation.

Motivated by this observation, we introduce \method, a unified loop-closure framework that reuses a single GFM backbone, MASt3R, for retrieval,
verification, and relative pose estimation. We attach a
SALAD aggregator~\cite{izquierdo2024salad} to the MASt3R encoder and train
the resulting global descriptors with contrastive learning. For verification,
we build on the pairwise reasoning paradigm of
Doppelgangers++~\cite{xiangli2025doppelgangers} to classify retrieved pairs from
intermediate MASt3R decoder states. Finally, we retain MASt3R's dense outputs
and convert them into relative camera constraints for the downstream pose-graph
optimization.

A key design principle of \method is that the three tasks do not merely
coexist as independent heads: the same image encoding is reused across
retrieval and pairwise decoding, avoiding redundant backbone inference.
Training is staged rather than fully joint---we first train the SALAD
retrieval branch, then freeze it and supervise the verification head on
hard negatives mined from the trained retrieval branch.

Our main contributions are summarized as follows:
\begin{itemize}[leftmargin=*,nosep]
    \item We propose \method, a unified loop-closure detection framework
    that maps the hierarchical representations of a 3D geometric
    foundation model onto the retrieval, verification, and relative pose
    estimation stages, sharing a single backbone across all three.

    \item We instantiate this design through (i) a MASt3R-encoder
    retrieval branch that aggregates encoder tokens into compact global
    place descriptors without a separate image backbone, (ii) a
    lightweight verification transformer that classifies retrieved pairs
    from pair-conditioned MASt3R decoder states, and (iii) the retained
    MASt3R geometric predictions, converted into relative camera
    constraints for pose-graph optimization.

    \item We establish a staged training and evaluation protocol that
    measures both module-level performance and end-to-end
    retrieval-conditioned loop-closure quality.
\end{itemize}
The rest of the paper is organized as follows.
\Cref{sec:related} reviews related work. \Cref{sec:method} presents the
\method framework and training strategy, and \cref{sec:experiments} reports
the experimental evaluation. Finally, \cref{sec:conclusion} concludes the
paper and outlines future work.
\begin{figure*}[ht]
    \centering
    \includegraphics[width=0.9\linewidth]{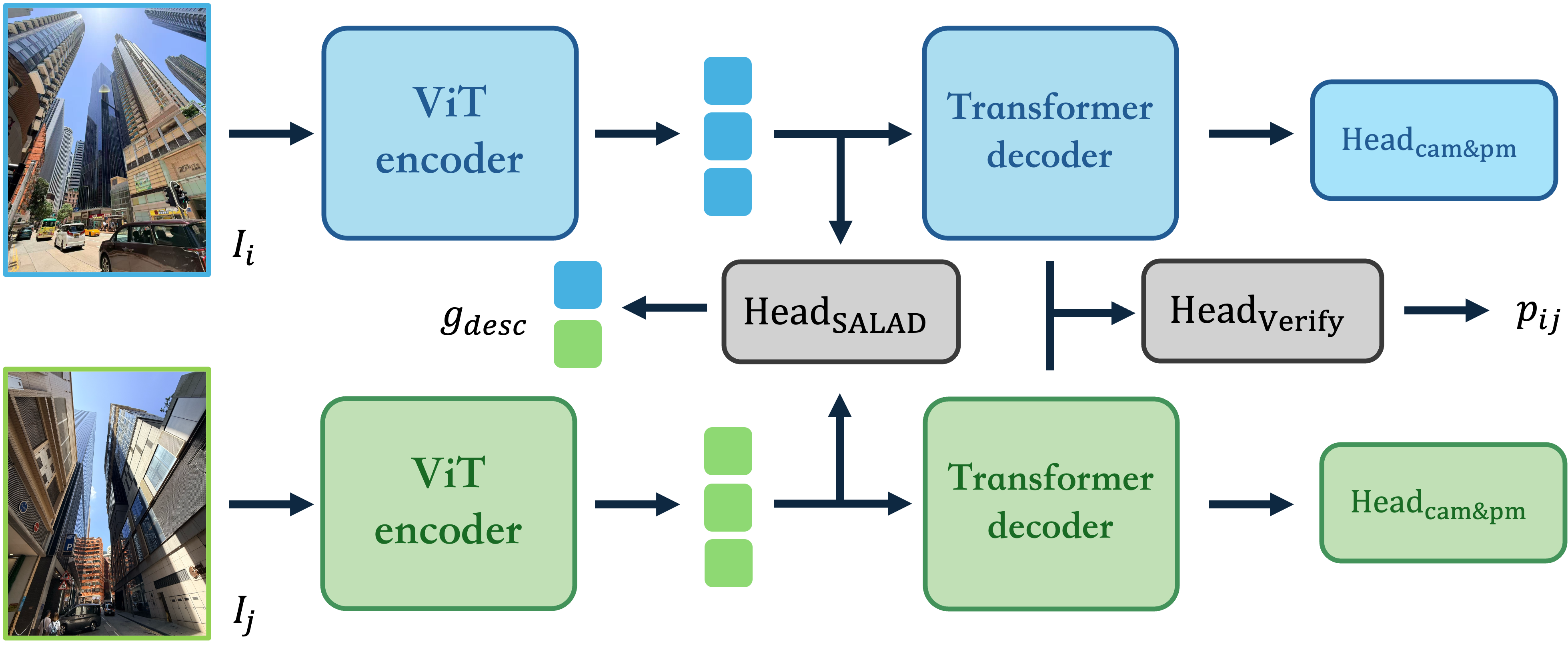}
    \caption{\textbf{Architecture of UniLoop} for a candidate image pair ($I_i$,$I_j$). A shared-weight MASt3R ViT encoder processes each image once. The SALAD head aggregates the resulting single-view tokens into normalized global descriptors $\mathbf{g}_i$ and $\mathbf{g}_j$ for candidate retrieval. For each shortlisted pair, the same tokens are passed to the cross-view decoder to obtain pair-conditioned features. A lightweight verification head predicts the loop probability $p_{ij}^{\mathrm{loop}}$, while the retained MASt3R's camera and pointmap heads recover the geometry used to construct relative-pose constraints for pose-graph optimization. Blue and green denote the two input views and their corresponding feature streams.
}
    \label{fig:framework}
\end{figure*}
\section{Related Work}
\label{sec:related}

\subsection{Visual Place Recognition}

Modern visual place recognition (VPR) aggregates patch-level features from
pretrained vision transformers into compact global descriptors for retrieval
under viewpoint and appearance changes~\cite{ye2023condition,oquab2023dinov2}.
SALAD~\cite{izquierdo2024salad} uses optimal-transport aggregation with a
dustbin to suppress non-informative regions, and
MegaLoc~\cite{berton2025megaloc} scales this paradigm through large-scale
training, but their representations are independent of geometric downstream
tasks. \method instead applies SALAD-style aggregation to the
MASt3R~\cite{leroy2024mast3r} encoder, so the same geometry-aware tokens
support both retrieval and subsequent geometric reasoning.

\subsection{Loop Closure Detection and Geometric Verification}

Loop closure detection typically retrieves candidates before testing their
geometric consistency~\cite{yu2024gv}. Because global descriptors are prone to
perceptual aliasing~\cite{yu2026rover}, verification commonly relies on local
correspondences and epipolar geometry or learned pairwise classifiers.
Doppelgangers++~\cite{xiangli2025doppelgangers} demonstrates that geometric features from GFMs can disambiguate image pairs. \method builds on this idea by
classifying intermediate, pair-conditioned decoder states from the shared
backbone, thereby rejecting geometrically inconsistent retrieved candidates.

\subsection{Geometric Foundation Models}

Geometric foundation models (GFMs) predict dense, view-consistent 3D
representations from images. DUSt3R~\cite{dust3r_cvpr24} predicts pointmaps
for unconstrained pairs, and MASt3R~\cite{leroy2024mast3r} adds dense local
descriptors and reciprocal matching. VGGT~\cite{wang2025vggt} extends this
to multi-view reconstruction, and VGGT-Omega~\cite{wang2026vggt} scales it
with substantially larger training data. Their encoder--decoder--head
hierarchy naturally supports the retrieval, verification, and pose-estimation
tasks addressed by \method.

\subsection{Geometric Foundation Models for Loop Closure}
GFMs support navigation~\cite{zheng2026enhancing},
SLAM~\cite{maggio2025vggt-slam}, and crowd-sourced
mapping~\cite{jiao2026opennavmap}, but in loop closure they serve mainly
as frozen matchers or pose regressors:
FLOCC-SLAM~\cite{lajoie20253d} pairs a VPR model with a frozen MASt3R pose
regressor; MASt3R-SLAM~\cite{murai2024_mast3rslam} reuses features for
retrieval and pointmap matching without training; and
MR.ScaleMaster~\cite{ju2026mr} though leverages various GFMs as front-end, it still relies on backend heuristics for loop
rejection and scale consistency.
\method instead trains task-specific retrieval and verification heads on
the MASt3R hierarchy, while retaining its geometric heads for relative
pose estimation.
\section{\method}
\label{sec:method}

\subsection{Problem Formulation and Overview}
\label{sec:problem}

Let a robot observe an RGB image stream $\mathcal{I} = \{I_t\}_{t=1}^{T}$, $I_t \in \R^{3 \times H \times W}$.
Given a query image $I_q$ and a database
$\mathcal{D}=\{I_j\}_{j=1}^{M}$, full loop closure must (i) retrieve a
compact candidate set $\mathcal{C}_q \subset \mathcal{D}$, (ii) decide
whether each pair $(I_q,I_j)$ is a valid loop, and (iii) estimate a
relative geometric constraint $\mathbf{T}_{qj}$ for each accepted pair.
We therefore seek
\begin{equation}
\mathcal{F}(I_q,I_j)
=
\left(
    s_{qj}^{\mathrm{ret}},
    p_{qj}^{\mathrm{loop}},
    \mathbf{T}_{qj},
    \boldsymbol{\Omega}_{qj}
\right),
\end{equation}
where $s_{qj}^{\mathrm{ret}}$ is the global retrieval similarity,
$p_{qj}^{\mathrm{loop}}$ the verification probability,
$\mathbf{T}_{qj}\in \SE(3)$ the relative transformation, and
$\boldsymbol{\Omega}_{qj}$ an optional information/confidence term for
pose graph optimization.
\method instantiates $\mathcal{F}$ by reusing a single geometric
foundation model across all three outputs, detailed in \cref{fig:framework}, which illustrates the framework.
A shared MASt3R encoder extracts patch tokens, a SALAD retrieval head
converts them into a global descriptor for candidate retrieval.
For each retrieved pair, the same encoder outputs feed the cross-view
decoder, whose intermediate features a lightweight verification
transformer classifies into a loop candidate probability.
The pair model's geometric outputs are retained for relative pose
estimation.

\subsection{Global Retrieval Branch}
\label{sec:encoder}

\paragraph{Shared single-image encoder}

Given image $I_i$, the MASt3R encoder produces patch tokens $\mathbf{F}_i
=
\encoder(I_i)
\in
\R^{N_i \times C_e},$
where $N_i$ is the number of image patches and $C_e$ is the encoder
dimension.
For the ViT-L MASt3R encoder used in our implementation,
$C_e=1024$.
The tokens are reshaped into a spatial feature map $\widetilde{\mathbf{F}}_i
\in
\R^{C_e \times H_p \times W_p}$.

The central computational constraint of \method is that
$\mathbf{F}_i$ is computed once per image and reused by both retrieval
and pairwise decoding.
This distinguishes the proposed framework from a naive ensemble in
which separate retrieval and verification networks independently encode
the same RGB input.

\paragraph{Global retrieval from encoder tokens}
\label{sec:retrieval}

We attach a SALAD-style optimal-transport aggregator
$\retriever_{\phi}$ to the MASt3R encoder features.
Because MASt3R does not provide a dedicated CLS token, we construct an
initial global token by mean pooling:
\begin{equation}
\mathbf{t}_i
=
\frac{1}{N_i}
\sum_{n=1}^{N_i}
\mathbf{F}_{i,n}.
\end{equation}

The retrieval descriptor is:
\begin{equation}
\mathbf{g}_i
=
\operatorname{norm}
\left[
    \retriever_{\phi}
    \left(
        \widetilde{\mathbf{F}}_i,
        \mathbf{t}_i
    \right)
\right]
\in \R^{D_g},
\end{equation}
where $\operatorname{norm}(\cdot)$ denotes $\ell_2$ normalization.
Using $K=64$ clusters, cluster dimension $D_c=128$, and token
dimension $D_t=256$ gives $D_g = K D_c + D_t = 8448$.
Retrieval similarity between query $q$ and database image $j$ is $s_{qj}^{\mathrm{ret}}
=
\mathbf{g}_q^\top \mathbf{g}_j$.
The top-$K_r$ images according to
$s_{qj}^{\mathrm{ret}}$ define the candidate set
$\mathcal{C}_q$.

\subsection{Verification Branch}
\label{sec:decoder}

\paragraph{Pair-conditioned cross-view decoder}

For a candidate pair $(I_i,I_j)$, the single-image tokens are passed to
the pretrained MASt3R cross-view decoder:
\begin{equation}
\left\{
\mathbf{H}_{ij}^{(\ell)},
\mathbf{H}_{ji}^{(\ell)}
\right\}_{\ell=1}^{L}
=
\decoder_{\psi}
\left(
    \mathbf{F}_i,
    \mathbf{F}_j
\right).
\end{equation}
Unlike the encoder tokens, the decoder features are pair-conditioned:
each representation is constructed in the context of the other image.
This makes them suitable for reasoning about whether the two views can
be reconciled geometrically.

Following the existing Doppelgangers++~\cite{xiangli2025doppelgangers} design, we extract a configurable
set of hidden layers, i.e., encoder/decoder stages corresponding to $\{0,6,9,12\}$ in the default 12-layer decoder configuration.
We denote the resulting token sequence by
\begin{equation}
\mathbf{Z}_{ij}
=
\operatorname{BuildTokens}
\left(
    \{
    \mathbf{H}_{ij}^{(\ell)}
    \}_{\ell\in\mathcal{S}},
    \{
    \mathbf{H}_{ji}^{(\ell)}
    \}_{\ell\in\mathcal{S}}
\right).
\end{equation}

\paragraph{Loop verification and relative pose constraints}
\label{sec:verification}
The verification transformer reasons over the decoder tokens where a lightweight transformer encoder produces $\mathbf{Y}_{ij}
=
\verifier_{\omega}^{\mathrm{tr}}
\left(
    \mathbf{Z}_{ij}
\right)$.
After pooling, a binary classifier predicts
\begin{equation}
\mathbf{z}_{ij}^{\mathrm{loop}}
=
\verifier_{\omega}^{\mathrm{cls}}
\left(
    \operatorname{Pool}
    (\mathbf{Y}_{ij})
\right)
\in \R^2,
\end{equation}
and
\begin{equation}
p_{ij}^{\mathrm{loop}}
=
\operatorname{softmax}
\left(
    \mathbf{z}_{ij}^{\mathrm{loop}}
\right)_1.
\end{equation}

The pairwise MASt3R branch additionally predicts dense geometric
outputs, including pointmaps, confidence maps, and local matching
features.
We denote these outputs by $\mathbf{G}_{ij}
=
\mathcal{G}
\left(
    \mathbf{H}_{ij}^{(L)},
    \mathbf{H}_{ji}^{(L)}
\right).$
A relative pose module converts the predicted pairwise geometry into $\left(
    \mathbf{T}_{ij},
    c_{ij}^{\mathrm{pose}},
    \boldsymbol{\Omega}_{ij}
\right)
=
\posehead(\mathbf{G}_{ij}),$
where
$\mathbf{T}_{ij}\in\SE(3)$ is the transformation from camera $i$ to
camera $j$.
An accepted loop edge is:
\begin{equation}
e_{ij}
=
\left(
    i,
    j,
    \mathbf{T}_{ij},
    \boldsymbol{\Omega}_{ij}
\right)
\quad
\text{if}
\quad
p_{ij}^{\mathrm{loop}} \ge \tau_{\mathrm{loop}}.
\end{equation}
For a query image $I_q$, the complete inference proceeds as in Algorithm 1.

\begin{algorithm}[t]
\caption{\method inference}
\label{alg:uniloop}
\begin{algorithmic}[1]
\Require Query $I_q$, database $\mathcal{D}$, top-$K_r$
\State Encode $I_q$ once:
$\mathbf{F}_q \gets \encoder(I_q)$
\State Compute descriptor:
$\mathbf{g}_q \gets \retriever(\mathbf{F}_q)$
\State Retrieve candidate set
$\mathcal{C}_q \gets \operatorname{TopK}
(\mathbf{g}_q,\mathcal{D})$
\For{$I_j \in \mathcal{C}_q$}
    \State Reuse cached $\mathbf{F}_j,\mathbf{g}_j$
    \State Decode pair:
    $\mathbf{H}_{qj}\gets\decoder(\mathbf{F}_q,\mathbf{F}_j)$
    \State Verify:
    $p_{qj}^{\mathrm{loop}}
    \gets\verifier(\mathbf{H}_{qj})$
    \If{$p_{qj}^{\mathrm{loop}}\ge\tau_{\mathrm{loop}}$}
        \State Estimate
        $\mathbf{T}_{qj},\boldsymbol{\Omega}_{qj}$
        from geometric output
        \State Add loop edge
        $(q,j,\mathbf{T}_{qj},\boldsymbol{\Omega}_{qj})$
    \EndIf
\EndFor
\end{algorithmic}
\end{algorithm}

\subsection{Training Strategy}
We use staged optimization to avoid destabilizing the pretrained geometric
backbone and to independently study the contribution of each branch. During retrieval
adaptation, we distill knowledge from a frozen MegaLoc~\cite{berton2025megaloc}
teacher into our
MASt3R--SALAD retrieval branch. The teacher is used only during training and
therefore introduces no inference-time cost.

\subsubsection{Stage I: Retrieval Adaptation on GSV-Cities}

Given a batch containing multiple images per place, the shared MASt3R
encoder and SALAD head produce normalized global descriptors
$\{\mathbf{g}_i\}$. We first optimize the Multi-Similarity metric-learning
loss with hard-pair mining:
\begin{equation}
\mathcal{L}_{\mathrm{ret}}
=
\mathcal{L}_{\mathrm{MS}}
\left(
    \{\mathbf{g}_i\},
    \{y_i^{\mathrm{place}}\}
\right),
\end{equation}
where $\mathcal{L}_{\mathrm{MS}}$ denotes Multi-Similarity loss with
hard-pair mining.

To transfer the teacher's retrieval knowledge, the frozen MegaLoc model
produces normalized descriptors $\{\mathbf{g}^{\mathrm{T}}_i\}$ for the same
images. The per-image distillation term $\mathcal{L}_{\mathrm{desc}}$
maximizes the cosine similarity between the student and teacher descriptors
extracted from each identical image:
\begin{equation}
\mathcal{L}_{\mathrm{desc}}
=
\frac{1}{B}\sum_i
\left(1-\mathbf{g}_i^{\top}\mathbf{g}^{\mathrm{T}}_i\right).
\end{equation}

The resulting Stage-I objective is
\begin{equation}
\begin{aligned}
\mathcal{L}_{\mathrm{stage\text{-}I}}
={}& \mathcal{L}_{\mathrm{MS}} \\
&{}+ \lambda_{\mathrm{desc}}\mathcal{L}_{\mathrm{desc}},
\end{aligned}
\end{equation}
where $\lambda_{\mathrm{desc}}$ balances the distillation term. This
objective retains place-level supervision while encouraging the student to
match the MegaLoc descriptors.
Earlier MASt3R encoder blocks, the cross-view decoder, and geometric
heads remain frozen.

\subsubsection{Stage II: Retrieval-Conditioned Pair Mining on MSLS}

After Stage I, the retrieval branch is frozen and used to mine
verification candidates.
For each MSLS query $I_q$, we retrieve top-$K_r$ database images $\mathcal{C}_q
=
\operatorname{TopK}
\left(
    \mathbf{g}_q^\top\mathbf{g}_j
\right).$
The MSLS annotations label retrieved pairs as positive, negative, or ignored.
Rather than sampling negatives uniformly at random, we preferentially retain
high-ranking retrieved negative pairs as hard negatives. These
retrieval-conditioned negatives are visually similar but nonmatching
candidates produced by the Stage-I retriever, and train the verifier to reject
the difficult false positives it encounters at inference time.
\subsubsection{Stage III: Verification Training}
For each mined pair $(I_i,I_j,y_{ij})$, the encoder, retrieval head,
MASt3R decoder, and geometry heads are initially frozen.
We train only the verification transformer $\verifier_{\omega}$.
The binary verification loss is:
\begin{equation}
\mathcal{L}_{\mathrm{ver}}
=
-\alpha_{y_{ij}}
\log
p(y_{ij}\mid I_i,I_j),
\end{equation}
where $\alpha_{y}$ optionally compensates class imbalance.
Focal loss can be substituted when hard negatives dominate the
training distribution.
\section{Experiments}
\label{sec:experiments}

We evaluate the global retrieval and loop-verification branches separately.
Retrieval is assessed with Recall@$K$ ($K\in\{1,5,10\}$), i.e., the fraction
of queries whose ground-truth match is retrieved among the top-$K$ database
images. For loop verification, we report average precision (AP), ROC-AUC, and
maximum recall at 100\% precision (MR). This branch-wise evaluation isolates
the contribution of candidate retrieval and subsequent geometric
disambiguation.

\subsection{Datasets}
\label{ssec:datasets}

\paragraph{Training.}
Stage-I retrieval adaptation is trained on
GSV-Cities~\cite{ali2022gsv}, which provides multiple street-level views
per place and thus supports per-place metric-learning batches with
natural positive pairs.
Stage-II pair mining and Stage-III verification training use
Mapillary Street-Level Sequences (MSLS)~\cite{Warburg_2020_msls}, a
large-scale multi-city dataset with substantial appearance, viewpoint,
and seasonal change.

\paragraph{Evaluation.}
We evaluate retrieval on five standard VPR benchmarks: MSLS~\cite{Warburg_2020_msls},
Pitts250k and Pitts30k~\cite{Gronat_2013_cvpr_pitts,Arandjelovic_2018_netvlad},
Tokyo~24/7~\cite{Torii_2018_tokyo247}, and Baidu~\cite{Sun_2017_Baidu_dataset}.
Together, these benchmarks cover city-scale retrieval under substantial
viewpoint and appearance variation.
For verification, we evaluate on the MSLS validation set, which contains 150K positive pairs and 180K negative pairs for loop verification.

\subsection{Comparison with State of the Art}
\label{ssec:sota}

\begin{table*}[t]
\centering
\caption{
\textbf{Recall@1 and Recall@10 on multiple VPR datasets.}
The best, second-best, and third-best results in each metric are highlighted
in dark, medium, and light green, respectively.
}
\begin{adjustbox}{width=0.99\linewidth}
\begin{tabular}{llc*{5}{cc}}
\toprule
\multirow{2}{*}{Method}
& \multirow{2}{*}{Venue}
& \multirow{2}{*}{Dim.}
& \multicolumn{2}{c}{MSLS val \cite{Warburg_2020_msls}}
& \multicolumn{2}{c}{Pitts250k \cite{Gronat_2013_cvpr_pitts,Arandjelovic_2018_netvlad}}
& \multicolumn{2}{c}{Pitts30k \cite{Gronat_2013_cvpr_pitts,Arandjelovic_2018_netvlad}}
& \multicolumn{2}{c}{Tokyo 24/7 \cite{Torii_2018_tokyo247}}
& \multicolumn{2}{c}{Baidu \cite{Sun_2017_Baidu_dataset}} \\
&
&
& R1 & R10
& R1 & R10
& R1 & R10
& R1 & R10
& R1 & R10 \\
\midrule

NetVLAD \cite{Arandjelovic_2018_netvlad}
& TPAMI~'18
& 4096
& 54.5 & 70.4
& 85.9 & 95.0
& 85.0 & 94.4
& 69.8 & 82.9
& 69.0 & \thirdresult{95.0} \\

AP-GeM \cite{Revaud_2019_ap_gem}
& ICCV~'19
& 2048
& 56.0 & 72.9
& 80.0 & 93.5
& 80.7 & 94.1
& 57.5 & 77.5
& 59.8 & 90.8 \\

CosPlace \cite{Berton_2022_cosPlace}
& CVPR~'22
& 2048
& 85.0 & 92.6
& 92.3 & 98.4
& 90.9 & 96.7
& 87.3 & 95.6
& 52.0 & 80.4 \\

MixVPR \cite{Alibey_2023_mixvpr}
& WACV~'23
& 4096
& 83.2 & 91.9
& 94.3 & \thirdresult{98.9}
& 91.6 & 96.4
& 87.0 & 94.0
& 71.9 & 94.7 \\

EigenPlaces \cite{Berton_2023_EigenPlaces}
& ICCV~'23
& 2048
& 85.9 & 93.1
& 94.1 & 98.7
& 92.5 & 97.6
& 93.0 & 97.5
& 69.1 & 91.9 \\

AnyLoc \cite{keetha2023anyloc}
& RAL~'23
& 49152
& 58.7 & 74.5
& 89.4 & 98.0
& 86.3 & 96.7
& 87.6 & 97.5
& \secondresult{75.6} & \secondresult{95.2} \\

SALAD \cite{Izquierdo_2024_SALAD}
& CVPR~'24
& 8448
& 88.2 & 95.0
& \thirdresult{95.0} & \secondresult{99.2}
& 92.3 & 97.4
& \thirdresult{94.6} & \secondresult{98.1}
& 72.7 & 93.6 \\

CricaVPR \cite{Lu_2024_cricavpr}
& CVPR~'24
& 10752
& 76.7 & 87.2
& 92.6 & 98.3
& 90.0 & 96.7
& 82.9 & 93.7
& 65.6 & 93.2 \\

CliqueMining \cite{Izquierdo_2024_cliqueM}
& ECCV~'24
& 8448
& \bestresult{91.6} & \secondresult{95.9}
& \secondresult{95.3} & \secondresult{99.2}
& \thirdresult{92.6} & \thirdresult{97.8}
& \bestresult{96.8} & \thirdresult{97.8}
& \thirdresult{72.9} & 92.7 \\

MegaLoc~\cite{berton2025megaloc}
& CVPRW~'25
& 8448
& \secondresult{91.0} & \thirdresult{95.8}
& \bestresult{96.4} & \bestresult{99.3}
& \bestresult{94.1} & \secondresult{98.2}
& \secondresult{96.5} & \bestresult{99.4}
& \bestresult{87.7} & \bestresult{98.0} \\

UniLoop (Ours)
& --
& 8448
& \thirdresult{90.1} & \bestresult{96.6}
& 93.4 & 97.3
& \secondresult{93.6} & \bestresult{99.3}
& 92.1 & 96.3
& 64.0 & 78.1
\\
\bottomrule
\end{tabular}
\end{adjustbox}
\label{tab:vpr}
\end{table*}

\begin{table*}[t]
    \centering
    \caption{Retrieval branch ablation on MSLS and Pitts30k-val, and Pitts30k-test.
    ``Blks'' is the number of fine-tuned MASt3R encoder blocks;
    ``Dist.'' indicates whether distillation is used
    (\cmark/\xmark). ``Avg. Rel. Imp.'' is the mean relative improvement
    across the nine reported metrics with respect to the first row.}
    \label{tab:abl_retrieval}
    \resizebox{\linewidth}{!}{
    \begin{tabular}{c c c c ccc ccc ccc}
        \toprule
        \multirow{2}{*}{Trainable Blks} & \multirow{2}{*}{Distill.} & \multirow{2}{*}{Resolution} & \multirow{2}{*}{\shortstack{Avg. Rel.\\Imp. (\%)}} &
        \multicolumn{3}{c}{MSLS} & \multicolumn{3}{c}{Pitts30k-val} & \multicolumn{3}{c}{Pitts30k-test} \\
        \cmidrule(lr){5-7}\cmidrule(lr){8-10}\cmidrule(lr){11-13}
        & & & & R@1 & R@5 & R@10 & R@1 & R@5 & R@10 & R@1 & R@5 & R@10  \\
        \midrule
        0 & \xmark & 224/224 & -- & 76.35 &84.59 & 87.97 & 90.17 &97.24 & 98.26 &88.13 &94.06 &95.58 \\
        2 & \xmark & 224/224 & 3.09 & 83.65 & 89.86 & 92.55  & 92.55 & 98.05 & 98.82 & 89.60 & 94.63 & 96.02 \\
        2 & \cmark & 224/224 & 3.75 & 83.65 & 91.22 & 93.65  & 93.48 & 98.38 & 99.07 & 90.18 & 95.03 & 96.45  \\
        2 & \xmark & 224/512 & 4.29 & 87.16 & 93.24  & 93.92 & 92.18& 97.66 & 98.62 & 90.17 & \ul{95.30} & 96.33  \\
        4 & \xmark & 224/224 & 3.50 & 84.32 & 90.95 & 92.84 & 93.20 & 97.95 & 98.70 & 90.05 & 94.76 & 96.07    \\
        4 & \cmark & 224/224 & 4.69 & 85.54 &93.65 &95.41 & \tb{93.73} &98.55 &\tb{99.30} & \ul{90.35} &95.13 &96.57  \\
        4 & \cmark & 224/512 & 5.42 & \ul{89.86}  &\ul{95.00}  &
        \ul{95.81} & 93.28  & \ul{98.55} &99.19 & 89.70  &95.09  &\tb{96.74}  \\
        4 & \cmark & 336/512 & \tb{5.87} & \tb{90.14} & \tb{95.54} & \tb{96.62} & \ul{93.60} & \tb{98.57} & \ul{99.28} & \tb{90.67} & \tb{95.47} & \tb{96.88} \\
        \bottomrule
    \end{tabular}
    }
\end{table*}

\subsubsection{Retrieval Branch}
Table~\ref{tab:vpr} compares \method with classical VPR baselines,
including NetVLAD~\cite{Arandjelovic_2018_netvlad} and
AP-GeM~\cite{Revaud_2019_ap_gem}, and recent learned methods, including
CosPlace~\cite{Berton_2022_cosPlace}, MixVPR~\cite{Alibey_2023_mixvpr},
EigenPlaces~\cite{Berton_2023_EigenPlaces},
SALAD~\cite{Izquierdo_2024_SALAD}, CricaVPR~\cite{Lu_2024_cricavpr},
CliqueMining~\cite{Izquierdo_2024_cliqueM}, and
MegaLoc~\cite{berton2025megaloc}. \method achieves the best Recall@10 on
MSLS (96.6) and Pitts30k (99.3). It ranks third on MSLS Recall@1 (90.1) and
second on Pitts30k Recall@1 (93.6), while exceeding the MegaLoc teacher on
MSLS Recall@10 (96.6 versus 95.8). These results indicate that the shared
MASt3R representation retrieves correct places reliably within the candidate
set, which is particularly important for the subsequent verification stage.

On Pitts250k and Tokyo~24/7, \method obtains 93.4/97.3 and 92.1/96.3
Recall@1/10, respectively, but remains behind MegaLoc. Its performance on
Baidu is also lower than that of the strongest dedicated VPR methods. Thus,
although the proposed shared representation is competitive on MSLS and
Pitts30k, improving cross-domain generalization remains an important
direction for future work.

\subsubsection{Verification Branch}
Table~\ref{tab:verification} compares image-retrieval (IR), geometric
verification (GV), and neural-classifier (NC) baselines. The proposed
hard-negative-trained NC verifier achieves the strongest results among the
listed methods, with 95.63 AP, 96.29 ROC-AUC, and 0.52 MR. It outperforms
the GV baselines SP+LG and eLoFTR and improves over the IR-only SALAD score,
showing the value of pair-conditioned MASt3R decoder features for rejecting
difficult retrieved candidates.
However, the MR of 0.52 shows that the NC branch does not yet retain all
true loops at a strict 100\%-precision operating point. This leaves room for
more powerful verification heads and end-to-end training.
\subsection{Ablation}
\label{ssec:abl_retrieval}

\begin{table}[t]
\centering
\caption{Comparison and ablation of the loop verification branch.
IR denotes image retrieval, GV denotes geometric verification, and NC denotes
the proposed neural classifier.}
\label{tab:verification}
\resizebox{0.95\linewidth}{!}{
\begin{tabular}{clccc}
\toprule
& Method &
AP &
ROC-AUC &
MR\\
\midrule
\multirow{2}{*}{\rotatebox{90}{IR}}
& SALAD & 94.85 & 94.97 & 0.25\\
& Ours (VPR-branch) & 92.41 & 92.88 & 0.31 \\
\midrule
\multirow{2}{*}{\rotatebox{90}{GV}}
& SP+LG & 84.67 & 89.87 & 0.0\\
& eLoFTR & 86.94 & 91.31 & 0.0 \\
\midrule
\multirow{5}{*}{\rotatebox{90}{NC}}
& Train from scratch & 93.22 & 94.36 & 0.12 \\
& \cornerarrow + Hard Negatives& 94.00 & \ul{95.49} & \ul{0.45}\\
& Pre-trained & 28.09 &7.66 & 0.0 \\
& \cornerarrow + Fine-tuned & \ul{94.23} & 95.12 & 0.0 \\
& \cornerarrow + Hard Negatives & \tb{95.63}  & \tb{96.29}  & \tb{0.52} \\
\bottomrule
\end{tabular}
}
\end{table}

Table~\ref{tab:abl_retrieval} isolates three design choices, fine-tuned
encoder blocks, teacher model distillation, and training resolution on MSLS val,
Pitts30k-val, and Pitts30k-test.

\paragraph{Encoder adaptation.}
With a frozen encoder, MSLS Recall@1 is 76.35. Fine-tuning the final two
encoder blocks raises it to 83.65, demonstrating the benefit of adapting
MASt3R features for global retrieval.

\paragraph{MegaLoc distillation.}
At $224{\times}224$, adding MegaLoc distillation to a four-block
configuration improves MSLS Recall@1 from $84.32$ to $85.54$ and
Pitts30k-test Recall@10 from $96.07$ to $96.57$. It also raises MSLS
Recall@10 from $92.84$ to $95.41$, indicating that the teacher improves
the ranking of correct matches beyond the first retrieved result.

\paragraph{Training resolution.}
Holding four blocks and distillation fixed, MSLS Recall@1 rises from
$85.54$ ($224{\times}224$) to $89.86$ ($224{\times}512$) to $90.14$
($336{\times}512$); Recall@10 reaches $96.62$ at the highest resolution.

\paragraph{Best configuration.}
Four trainable blocks, distillation, and $336{\times}512$ resolution
give the overall best result (MSLS $90.14/95.54/96.62$, Pitts30k-test
$90.67/95.47/96.88$), reported in Table~\ref{tab:vpr}.

\paragraph{Verification adaptation and hard negatives.}
Table~\ref{tab:verification} shows that fine-tuning the pretrained MASt3R
verifier raises AP from 28.09 to 94.23. Adding retrieval-conditioned hard
negatives further increases performance to 95.63 AP, 96.29 ROC-AUC, and
0.52 MR. Hard negatives also improve the scratch-trained model, but it remains
behind the fine-tuned counterpart (94.00 AP and 0.45 MR), supporting the use of
pretrained representations and hard-negative mining.
\section{Conclusion and Future Work}
\label{sec:conclusion}

We introduced \method, a unified framework for full loop closure
detection built on the hierarchical representations of GFM, i.e., MASt3R.
The shared image encoder supports global retrieval through a trainable
SALAD aggregation head, pair-conditioned decoder features support
loop verification, and dense geometric outputs provide the basis for relative pose constraints.
By reusing a common backbone, \method connects candidate generation and
geometric validation while avoiding redundant image encoding.

Several directions remain for future work. First, \method can be extended
to multi-view GFMs, which jointly reason over more than two images. Such
backbones could provide a richer multi-view representation and improve
efficiency by reducing the repeated pairwise decoding required by the
currently two-view MASt3R backbone.
Second, our current MegaLoc distillation supervises only the final global
descriptor. Incorporating consistency constraints on intermediate latent
representations may transfer richer retrieval knowledge and further improve
the student retrieval branch.
Finally, we will deploy \method in real robot localization systems and
evaluate its impact on long-term pose-graph optimization. This evaluation
will establish the practical effectiveness of the unified retrieval,
verification, and the relative-pose pipeline under real-world operating conditions.

\bibliographystyle{IEEEtran}
\bibliography{refs}

\end{document}